\documentclass{article}

\usepackage{microtype}
\usepackage{subcaption}
\usepackage{graphicx}
\usepackage{subcaption}
\usepackage{booktabs} 

\usepackage{subcaption}
\usepackage{hyperref}

\usepackage[accepted]{icml2025}

\usepackage{amsmath}
\usepackage{amssymb}
\usepackage{mathtools}
\usepackage{amsthm}

\usepackage[capitalize,noabbrev]{cleveref}

\theoremstyle{plain}
\newtheorem{theorem}{Theorem}[section]

\newtheorem{lemma}[theorem]{Lemma}
\newtheorem{corollary}[theorem]{Corollary}
\theoremstyle{definition}

\newtheorem{assumption}[theorem]{Assumption}
\theoremstyle{remark}

\usepackage[textsize=tiny]{todonotes}

\icmltitlerunning{DiLoCoX: A Low-Communication Large-Scale Training Framework for Decentralized Cluster}

\begin{document}

\twocolumn[
\icmltitle{DiLoCoX: A Low-Communication Large-Scale Training Framework for Decentralized Cluster}




\begin{icmlauthorlist}
\icmlauthor{Ji Qi}{cmss}
\icmlauthor{WenPeng Zhu}{cmss}
\icmlauthor{Li Li}{cmss}
\icmlauthor{Ming Wu}{gravitylab}
\icmlauthor{YingJun Wu}{cmss}
\icmlauthor{Wu He}{cmss}
\icmlauthor{Xun Gao}{cmss}
\icmlauthor{Jason Zeng}{gravitylab}
\icmlauthor{Michael Heinrich}{gravitylab}
\end{icmlauthorlist}

\icmlaffiliation{cmss}{China Mobile(Suzhou) Software Technology, JiangSu, China}
\icmlaffiliation{gravitylab}{Zero Gravity Labs}

\icmlcorrespondingauthor{Ji Qi}{qiji@cmss.chinamobile.com}

\icmlkeywords{Machine Learning, ICML}

\vskip 0.3in
]



\printAffiliationsAndNotice{}

\begin{abstract}
The distributed training of foundation models, particularly large language models (LLMs), demands a high level of communication. Consequently, it is highly dependent on a centralized cluster with fast and reliable interconnects. \textit{Can we conduct training on slow networks and thereby unleash the power of decentralized clusters when dealing with models exceeding 100 billion parameters?} In this paper, we propose DiLoCoX, a low-communication large-scale decentralized cluster training framework. It combines \textit{Pipeline Parallelism with Dual Optimizer Policy}, \textit{One-Step-Delay Overlap of Communication and Local Training}, and an \textit{Adaptive Gradient Compression Scheme}. This combination significantly improves the scale of parameters and the speed of model pre-training. We justify the benefits of one-step-delay overlap of communication and local training, as well as the adaptive gradient compression scheme, through a theoretical analysis of convergence. Empirically, we demonstrate that DiLoCoX is capable of pre-training a 107B foundation model over a 1Gbps network. Compared to vanilla AllReduce, DiLoCoX can achieve a 357x speedup in distributed training while maintaining negligible degradation in model convergence. To the best of our knowledge, this is the first decentralized training framework successfully applied to models with over 100 billion parameters.
\end{abstract} 

\section{Introduction}
LLMs have quickly become dominant in the field of AI due to their exceptional capabilities. Their effectiveness in areas such as automated dialogue generation, machine translation, content summarization, and recommendation systems clearly demonstrates their superior advantages \cite{zhao2023survey,wang2022pre,bommasani2021opportunities}. However, as efforts to improve the accuracy of these models continue, both their complexity and scale have grown exponentially. In recent years, the number of parameters has increased from billions \cite{devlin2018bert} to trillions \cite{fedus2022switch, du2022glam}, introducing significant challenges in model training. For example, pre-training a model with 175B \cite{brown2020language} parameters can require 12,000 GPUs operating for 118 days \cite{narayanan2021efficient}.

In distributed foundation model training, communication is the key bottleneck. As an example, fine-tuning GPT-J-6B over 10B tokens with 262K batch size across 4 machines (2 A100 GPUs each) demands 915.5 TB data communication in total training\cite{cocktailsgd}. As a result, the high-speed centralized cluster is currently the dominant solution for training foundation model\cite{rendle2016robust}.

Decentralized clusters usually have a significant amount of computing resources than centralized cluster, but the bandwidth between clusters is relatively slower, with only hundreds of Mbps to 1 Gbps in common. When using a training framework such as Megatron-LM\cite{shoeybi2019megatron} for distributed cluster training, the bandwidth between clusters often becomes a training bottleneck\cite{dai2024high, zhang2022mics}, resulting in the inability to fully utilize the computing resources of decentralized clusters for model training.

Recently, there has been an exciting collection of work focusing on the decentralized training of foundation models. By introducing outer Nesterov momentum optimizers, DiLoCo \cite{douillard2023diloco} can train a model with up to 400M parameters using only data parallelism and achieve even better performance than a fully synchronous model. OpenDiLoCo\cite{jaghouar2024opendiloco} is an open-source implementation of DiLoCo. With FP16 gradient compression and Hivemind optimized collective communication operators, it can train the 1.1B model over poorly connected networks with a bandwidth of hundreds of Mbps and a negligible degradation in convergence. CocktailSGD \cite{cocktailsgd} combines three distinct compression techniques, achieving 117× aggressive compression in fine-tuning LLMs with up to 20B parameters over a 500Mbps network.

However, despite these recent efforts, the scale of model parameters and communication bottleneck is still a challenge when pretraining over 100B parameter models. DiLoCo and OpenDiLoCo \cite{douillard2023diloco, jaghouar2024opendiloco} have comparable model convergence with the fully synchronous model. However, they only use data parallelism and do not support FSDP \cite{zhao2023pytorchfsdpexperiencesscaling} or deepspeed \cite{rajbhandari2020zeromemoryoptimizationstraining}, thus the VRAM of GPU capacity limits the scale of model parameters. Furthermore, DiLoCo and OpenDiloCo \cite{douillard2023diloco, jaghouar2024opendiloco} employ a pseudo-gradients synchronous mechanism. During the synchronization of pseudo-gradients, local training is in an idle state. When the model scale is large and network bandwidth is limited, the idle time for synchronizing pseudo-gradients becomes unacceptable. CocktailSGD \cite{cocktailsgd} uses pipeline parallelism to support 20B model training, but the compression ratio of data parallelism is aggressive up to 117 times and does not use local training like DiLoCo or OpenDiLoCo in decentralized clusters, which has a potential impact on model convergence.

In order to train models with a scale of more than 100B parameters on low-bandwidth decentralized clusters while having comparative model convergence, we have identified the following key challenges: 
1. Introduce model parallelism 
to address the limitation of VRAM which has to accommodate the whole model parameters.
2. The overlap between the synchronization of pseudo-gradients and local training to avoid the idleness of computing resources.
3. Design an efficient gradient compression algorithm and balance it with the number of local training steps to ensure the convergence of model training. 

To address the above challenges, we propose a low communication large-scale model training framework \textbf{DiLoCoX} for decentralized cluster. Experiments demonstrate that DiLoCoX can pre-train a 107B model and significantly hide communication overhead while ensuring model convergence on decentralized clusters with only 1Gbps network bandwidth. To the best of our knowledge, this is currently the largest-scale model for effective decentralized cluster training. 

The main contributions are as follows.
\begin{itemize}
    \item \textbf{Pipeline Parallelism with Dual Optimizer Policy} To address the limitation of model parameter scale caused by the lack of support for model parallelism in DiLoCo-like framework, we propose Pipeline Parallelism with Dual Optimizer Policy into the framework and successfully expand the model parameters over 100B.

    \item \textbf{One-Step-Delay Overlap of Communication and Local Training} We propose a one-step-delay overlap mechanism between the synchronization of pseudo-gradients and local training to avoid the idleness of computing resources, which significantly improves the efficiency of model training.
    
    \item  \textbf{Design of Adaptive Gradient Compression Algorithm:}
    By thoroughly analyze different compression scheme, we have designed an efficient AllReduce-compatible communication compression scheme. In order to ensure the convergence of the model and full overlap with local training, we propose an adaptive gradient compression algorithm based on this compress scheme to trade-off compress ratio and the number of local training steps to avoid aggressive compress which cause potential degradation on model convergence.
    
    \item  \textbf{Theoretical Analysis of Convergence and Extensive Experiments:} We justify the benefit of the one-step-delay overlap of communication and local training,  adaptive gradient compression scheme through a theoretical analysis of convergence. Empirically, we demonstrate that DiLoCoX is capable of pre-training a 107B foundation model over a 1Gbps network. Compared to vanilla AllReduce, DiLoCoX can achieve a 357x speedup in distributed training while maintaining negligible degradation in model convergence.
\end{itemize}

\section{Method}
\subsection{Problem Formulation}
In this paper, the primary emphasis is laid on the parallel configuration comprising data parallelism $D$ and model parallelism $M$. A total of $N = D \times M$ workers are involved. Each $worker_{i,j}$ sustains a local data source $\mathcal{D}_i$, where $i$ and $j$ are the DP and MP indices, respectively. In this local data source, a local loss function $f_i$ is defined. All parallel workers cooperate in order to minimize the objective function $f:\mathbb{R}^d \to \mathbb{R}$, that is, to determine the parameters of the target model $\hat{\theta} \in \mathbb{R}^d$ such that
$
\hat{\mathbf{\theta}} = \mathrm{argmin}_{\mathbf{\theta} \in \mathbb{R}^d} \left[ f(\mathbf{\theta}) = \frac{1}{D} \sum_{i = 1}^D \mathbb{E}_{\zeta \sim \mathcal{D}_i} f_i(\mathbf{\theta}; \zeta) \right]
$. 
Here, $\zeta$ represents the data sampled from each local data source. At iteration $t$, each $worker_{i,j}$ holds a fraction of the local model replica denoted as $\theta^{(i,j)}_t$ and computes its local gradient over the data sample $\zeta^{(i)}_t$. Subsequently, all workers communicate and compute the average gradient. Then, $worker_{i,j}$ updates its fraction of the local model as follows:
$
\theta_{t + 1}^{(i,j)} = \theta_t^{(i,j)} - \eta \sum_{i = 1}^D \nabla f_i (\theta_t^{(i,j)} ; \zeta_t^{(i)})
$
where $\eta$ is the learning rate.

\subsection{Pipeline Parallelism with Dual Optimizer Policy}
\label{sec:ppdpp}
DiLoCo \cite{douillard2023diloco} and OpenDiLoCo \cite{jaghouar2024opendiloco} establish a standard for synchronous LocalSGD in language modeling with outer Nesterov momentum optimizers. However, both of them do not support model parallelism. Consequently, the model parallelism $M$ is regarded as 1.

For the OpenDiLoCo framework \cite{jaghouar2024opendiloco}, each worker $i$ performs $H$ local updates utilizing an inner optimizer on their corresponding data shard $\mathcal{D}_i$. Subsequently, the first worker on the node that also holds the outer optimizer computes the parameter change (pseudo gradient) $\delta^{(i)}_t=\theta_{t - 1}-\theta^i_t$ and averages the pseudo-gradient across clusters. Finally, the first worker of the node's outer optimizer performs a step operation to update local parameters and broadcasts the updated parameters to the remaining workers of the node.

The OpenDiLoCo framework requires GPU VRAM that is capable of accommodating the complete model parameters and inner optimizer states. Additionally, the first worker on the node is also required to store the state of the outer optimizer. This leads to unbalanced VRAM usage and consequently restricts the further expansion of the scale of model parameters.

\begin{figure}[ht]
\vskip 0.2in
\begin{center}
\centerline{\includegraphics[width=\columnwidth]{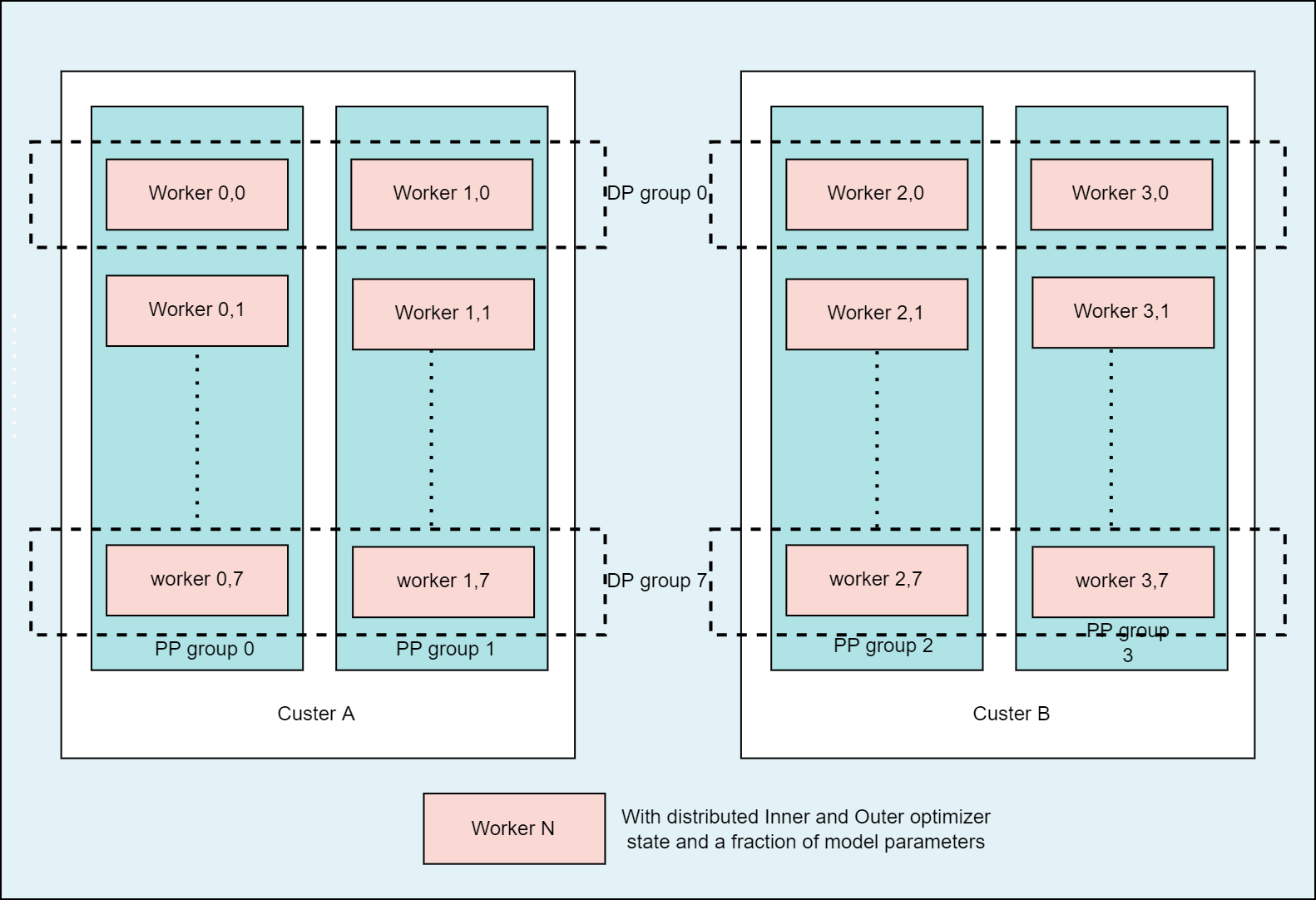}}
\caption{Pipeline Parallelism with Dual Optimizer. The example infrastructure comprises 32 workers distributed among 2 decentralized clusters, with 16 workers in each cluster. Each worker maintains distributed inner and outer optimizers' state and a fraction of model parameters. Two clusters are trained independently for $H$ steps respectively. The parallel strategy is PP = 8, DP = 2 for each cluster, and the parameters are updated by using their respective inner optimizers. Finally, the outer distributed optimizer performs a step operation to update local parameters.}
\label{fig:dilocox-framework}
\end{center}
\vskip -0.2in
\end{figure}

Based on the fact that pipeline parallelism is effective and has the least communication volume among parallel methods in the 3D parallel strategy \cite{narayanan2021efficient}, as shown in Figure \ref{fig:dilocox-framework}, we propose \textit{Pipeline Parallelism with Dual Optimizer Policy}. According to the pipeline parallelism approach, we divide the model into $M$ stages by layers and partition $N$ workers in decentralized clusters into $D$ data parallel groups, with the relationship $N = M×D$. Each $worker_{i,j}$ holds two distributed optimizers (inner and outer) and a fraction of the model parameters, and performs $H$ local updates using the inner optimizer on its corresponding data shard $\mathcal{D}_i$. Then, each worker computes its own parameter change (pseudo-gradient) and averages it with those of other workers in the same DP group. Finally, the distributed outer optimizer in each PP group performs a step operation to update the local parameters.

The advantage of Pipeline Parallelism with Dual Optimizer Policy resides in the fact that each worker merely stores a fraction of the model parameters, overcoming the limitation that the VRAM of a GPU must hold a complete model. Additionally, each worker incorporates a part of the state of the distributed outer optimizer, leading to a more balanced utilization of VRAM. Based on this scheme, we have implemented DiLoCoX training of models exceeding 100B parameters on decentralized clusters of NVIDIA 40G A800.

\subsection{One-Step-Delay Overlap of Communication and
Local Training}
Through introducing the \textit{Pipeline Parallelism with Dual Optimizer Policy} in section \ref{sec:ppdpp}, we can train the foundation model over 100B on decentralized clusters. However, DiLoCo and OpenDiLoCo \cite{douillard2023diloco, jaghouar2024opendiloco} adopt a synchronous approach between local training and pseudo-gradient averaging, which synchronizes the gradient of the outer optimizer after training $H$ steps. During the synchronization process, the computing resources are in idle state.

\begin{figure}[ht]
\vskip 0.1in
\begin{center}
\centerline{\includegraphics[width=\columnwidth]{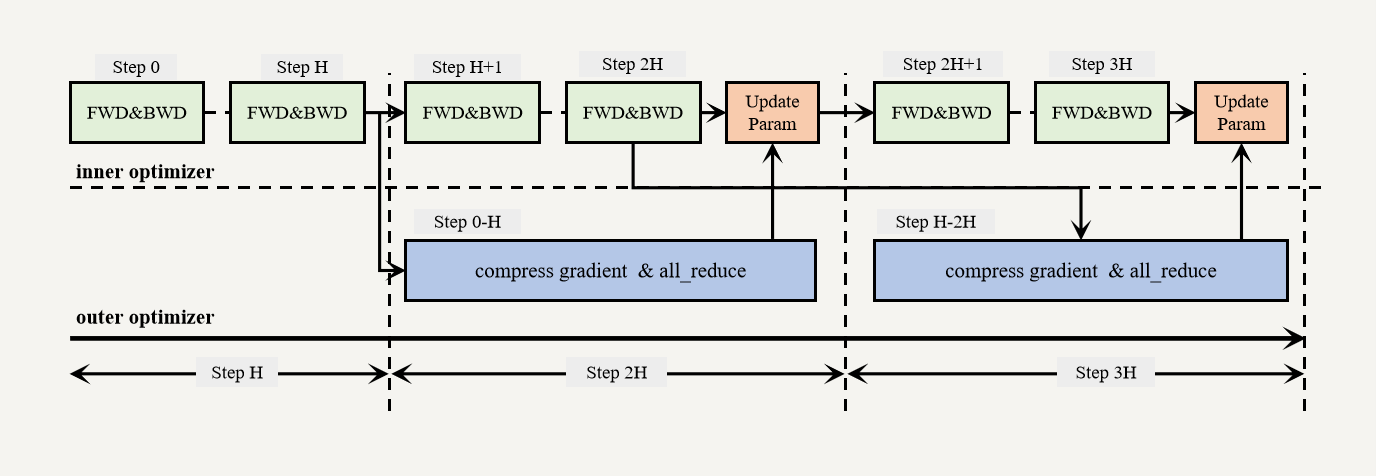}}
\caption{One-Step-Delay Overlap of Communication and Local Training.}
\label{fig:one-step-delay}
\end{center}
\vskip -0.3in
\end{figure}

We propose one-step-delay overlap of communication and local training illustrated in Figure \ref{fig:one-step-delay} with the following steps:
\begin{itemize}
\item After finishing the first global $H$-step, calculate current pseudo-gradients and start performing pseudo-gradients averaging asynchronously.
\item During the training of global steps between $H$ and $2H$, we average the last pseudo-gradients simultaneously using AllReduce.
\item After finishing the $2H$ step, we calculate current pseudo-gradients and start performing asynchronous averaging of current pseudo-gradients. Then we update the model parameter by the delayed last averaged pseudo-gradients.
\end{itemize}

In summary, the model parameters of outer step $t$ are obtained by $\theta^{t - 1}$ and $\triangle^{t - 1}$. It is denoted as $\theta^{(t)} \leftarrow OuterOpt(\theta^{(t - 1)},\triangle^{(t - 1)})$. We assume that the pseudo-gradients will not change significantly between two consecutive outer steps. Therefore, the overlap of communication and local training can greatly improve the model efficiency and will not have a significant impact on the convergence of the model. We verify this in subsequent experiments.

\subsection{Design of Adaptive Gradient Compression Algorithm}

\subsubsection{Analysis of Communication Overhead}
After $H$ steps of local training in $C$ clusters, we need to perform a pseudo-gradient update of decentralized clusters. Suppose that the number of model parameters is $\theta$, so the total number of parameters that must be communicated by all workers is $\theta \times D$, where $D$ is the degree of data parallelism.

In this paper, we solely take into account the communication overhead between clusters. Assuming the utilization of Ring AllReduce \cite{baidu2017ringallreduce} for pseudo-gradient updates, the total number of parameters that require communication between two clusters is $2\times(C - 1)\times\theta/C$.

Assuming that the pseudo-gradient is stored in FP32, the model parameter is 100B and there are three decentralized clusters, the communication overhead between the three clusters is approximately 533.3 GB for every $H$ step. If the bandwidth between decentralized clusters is 1 Gbps, transmitting 533.3 GB data would take 1.18 hours. Assuming that the local training step $H$ is 500 and the duration of every local step is 1 second, thus the total time of local training would take 0.13 hours. Even if we use section \ref{sec:ppdpp} \textit{Pipeline Parallelism with Dual Optimizer Policy}, the idle time of computing is approximately 1.04 hours, which is unacceptable for decentralized cluster training. For models with a scale of more than 100B, we need to design an effective compression algorithm to compress more than 10x, ensuring that the communication time of decentralized clusters is within a reasonable range.

\subsubsection{Design of Effective Compress Algorithm}
Currently, the major compression methods include sparsification \cite{Strom2015ScalableDD,wangni2017gradientsparsificationcommunicationefficientdistributed,alistarh2018convergencesparsifiedgradientmethods}, quantization\cite{alistarh2017qsgdcommunicationefficientsgdgradient} and Low-Rank\cite{vogels2020powersgdpracticallowrankgradient}, etc. These methods can significantly reduce the communication overhead of distributed training. However, none of these methods achieves a compression ratio larger than 10× without hurting the convergence\cite{cocktailsgd}.

CocktailSGD \cite{cocktailsgd} compresses communication aggressively using a combination of Top-K sparsification, Quantization, and Random sparsification and achieves up to 117× compression in fine-tuning LLMs up to 20B parameters. Aggressive compression leads to significant differences between the parameters of the local model and the global model, resulting in a degradation of the convergence of the global model, but the approach of combined different compression has inspired us to design an efficient compression algorithm. 

Firstly, we analyze four common compression algorithms: Random Sparsification, Top-K Compression, Quantization, and Low-Rank. 

\textbf{Random Sparsification} randomly selects a portion of the gradients for update according to the seed. By sending only a random seed, the sparsity pattern can be fully recovered. However, given the same sparsity ratio, random sparsification introduces more errors since it does not necessarily keep the values of the largest norm \cite{cocktailsgd}.

\textbf{Top-K Compression} selects the top-k elements with the largest values for communication and it has fewer compression errors (in l2 norm) compared to random sparsification. However, with $d$ numbers as input, top-k compression requires $K\log_{2}d$ bits (as index list) or $d$ bits (as bitmap). When $d$ is relatively large, it will lead to very large communication costs. In addition, Top-K compression is not AllReduce compatible and requires a parameter server and double compression, which is less efficient than AllReduce compatible compression algorithms which are essential for the 100B model \cite{cocktailsgd}.

\begin{algorithm}
   \caption{Compressor $\mathcal{C}[\delta]$}
   \label{alg:compressor}
\begin{algorithmic}
    \STATE {\bfseries Input:} Pseudo-gradient $\delta$, low-rank $r$, quantization $q$-bits.
    \STATE (1) Low-Rank $r$ Approximation:
    \STATE \hspace{3mm} $\delta_1 \leftarrow \text{LOWRANK}(\delta, r)$.
    \STATE (2) Quantize values to $q$ bits:
    \STATE \hspace{3mm} $\delta_2 \leftarrow \text{QUANTIZE}(\delta_1)$.
    \STATE {\bfseries Output:} Compressed pseudo-gradient $\delta_2$.
\end{algorithmic}
\vspace{-0.1cm} 
\end{algorithm}

 \textbf{Quantization} is efficient and compatible with AllReduce. However, if the original values are stored in FP16, it cannot achieve a compression ratio of more than 16 times. Furthermore, a linear decrease in the number of bits for quantization often results in an exponential increase in error. It is possible to apply quantization in conjunction with other compression algorithms that are compatible with AllReduce.

 \textbf{Low-Rank} argues that modern deep networks are over-parameterized and can use a Low-Rank update to update the gradient \cite{vogels2020powersgdpracticallowrankgradient}. It is sometimes possible to converge to the optimum using Low-Rank approximations of the gradients without the need for error feedback, and it can be implemented using the AllReduce, which is very efficient.

 \begin{algorithm}
    \caption{DiLoCoX Framework}
    \label{alg:dilocox_framework}
\begin{algorithmic}
    \STATE {\bfseries Input:} Initial Model $\theta^0$, number of workers $N$, Data Parallelism $D$, Pipeline Parallelism $M$, Data shards $\{\mathcal{D}_1,\cdots,\mathcal{D}_D\}$, error buffer $e$, distributed Optimizers $\text{InnerOpt}$ and $\text{OuterOpt}$, combined compress algorithm $\mathcal{C}$, gradient quantization $q$, gradient window $c$, initial local training step $H_1$ and initial Low-Rank compression rank $r_1$, adaptive gradient compress algorithm \textit{AdaGradCmp}. 
    \FOR{outer step $t=1 \cdots T $}
        \FOR {$i=1 \cdots D$}
            \FOR {$j=1 \cdots M$}
            \STATE // Start Local Training Thread
            \STATE $\theta^t_{i,j} \leftarrow \theta^{t-1}_{i,j}$
            \FOR{inner step $h=1 \cdots H_i $}
            \STATE $x \sim \mathcal{D}_i$
            \STATE $\mathcal{L} \leftarrow f(x, \theta^t_{i,j})$
            \STATE $\theta^t_{i,j} \leftarrow \text{InnerOpt}(\theta^t_{i,j}, \nabla_{\mathcal{L}}) $
            \ENDFOR
            \STATE // Start Compress and Communicate Thread
            \IF{$t$ \textgreater 1} 
            \STATE $\Delta^{t-1}_j \leftarrow \text{AllReduce}(\mathcal{C}(\delta^{t-1}_{i,j}|q, r_i))_{i \in \{1 \cdots D\}} $ 
            \STATE $e^{t}_{i,j} \leftarrow \delta^{t-1}_{i,j} - \Delta^{t-1}_j$
            \ENDIF
            \STATE // Waiting all threads to finish
            \STATE $r_{i+1}, H_{i+1} \leftarrow \text{AdaGradCmp}(c, r_i, H_i, \Delta^{t-1}_j) $
            \STATE $\delta^t_{i,j} \leftarrow (\theta^{(t-1)}_j -  \theta^t_{i,j}) + e^t_{i,j} $ // pseudo-gradients and error compensation
            \IF{$t$ \textgreater 1} 
            \STATE $\theta^t_j \leftarrow \text{OuterOpt}(\theta^{t-1}_{i,j}, \Delta^{t-1}_j) $
            \ENDIF
        \ENDFOR
        \ENDFOR
    \ENDFOR
\end{algorithmic}
\vspace{-0.1cm} 
\end{algorithm}

DiLoCoX is a variant of LocalSGD and DiLoCo. When updating the pseudo-gradient after local training for $H$ steps, we need to synchronize all gradient information as efficiently as possible to reduce the gradient differences between different nodes caused by compression. Therefore, we choose Quantization and Low-Rank compression methods that support AllReduce. To maximize the compression ratio, inspired by the combined different compression of CocktailSGD, we adopt a combined compression method of \textit{Quantization} and \textit{Low-Rank} which is illustrated in Algorithm \ref{alg:compressor}.

\subsubsection{Adaptive Gradient Compress Algorithm}
\textit{How to coordinate the number of steps of local training and two combined compression algorithms to balance the training efficiency and the convergence of the model when pre-training over a 100B model?} 

\begin{theorem}[Principle of Rank Diminishing \cite{feng2022rank}]
\label{thm:rank-diminishing}
Suppose that each layer \( f_i, i = 1, \dots, L \) of network \( F \) is almost everywhere smooth and data domain \( \mathcal{X} \) is a manifold, then both the rank of sub-networks and intrinsic dimension of feature manifolds decrease monotonically by depth:
\begin{align*}
    \operatorname{Rank}(f_1) &\geq \operatorname{Rank}(f_2 \circ f_1) \geq \cdots \\ & \geq  \operatorname{Rank}(f_{L-1} \circ \cdots \circ f_1) \geq \operatorname{Rank}(F_L), \\
    \dim(\mathcal{X}) &\geq \dim(\mathcal{F}_1) \geq \dim(\mathcal{F}_2) \geq \cdots \geq \dim(\mathcal{F}_L). 
\end{align*}
\end{theorem}

This principle \cite{feng2022rank} describes the behavior of generic neural networks with almost everywhere smooth components, which exhibits the monotonic decreasing (but not strictly) of network ranks and intrinsic dimensionality of feature manifolds. This property will affect the gradient rank of the parameter space through back-propagation. The direction of parameter updates in the network gradually focuses on a few principal components, increasing the redundancy of the gradient matrix and causing a natural decrease in rank.

This implies that as training progresses, the effective information of gradients gradually concentrates within a low-rank space. Therefore, during the model training process, as iterations deepen, we can progressively utilize lower ranks to further compress communication data.

\begin{algorithm}
   \caption{Adaptive Gradient Compress Algorithm}
   \label{alg:adaptiveGradientCompress}
\begin{algorithmic}
    \STATE {\bfseries Input:} Set the gradient rank window $c$, initialize the rank for practical compression $r_1$ and corresponding local training step $H_1$. Here $t$ denotes the iteration of outer optimizer, rank reduction ratio $\alpha_{r'}$.
        \STATE // After each outer optimizer completes the \texttt{AllReduce} operation.
        \STATE // Calculate \( r'_t \) with the globally averaged gradient.
        \IF{$t<c$}
            \STATE $r_t = r_1$ \\
            \STATE $\alpha_{r'} = 1$ \\
            \STATE $H_t = H_1$
        \ELSE
            \STATE $r_t = \frac{1}{c}\sum_{t-c+1}^{t}{r'_i}$ \\
            \STATE $\alpha_{r'} = \frac{r_1-r_{t}}{r_1}$
            \STATE $H_t = H_1\cdot\alpha_{r'}$
        \ENDIF
    \STATE {\bfseries Output:} Adaptive $r_t$ and $H_t$.
\end{algorithmic}
\vspace{-0.1cm} 
\end{algorithm}

On the other hand, Local SGD and quantization also provide some data compression capabilities. The overall compression effectiveness depends on their combined usage.

When the rank $r$ is dynamically and adaptively adjusted during the training process, the hyperparameter $H$ in Local SGD can also be continuously tuned accordingly to maximize the overlap between local training and gradient updates while ensuring model convergence.

We propose \textit{Adaptive Gradient Compress Algorithm} illustrated in Alg \ref{alg:adaptiveGradientCompress}. During the initial and intermediate phases of model training, gradients descend rapidly, leading to a sharp decrease in \( r'_{t} \). Our \( r_t \) dynamically responds to the changes in gradient ranks, enabling more efficient low-rank compression. And $H_t$ will also be adaptively adjusted accordingly by computing $\alpha_{r'}$. In the final phase, gradient descent slows significantly, and the low-rank parameters evolve at a reduced pace. Consequently, this leads to a stable $r_t$ and $H_t$, which means a stable low-rank compression.

\section{Theoretical Analysis}
\label{sec:theo}
\begin{assumption}
(Smoothness) For any $i \in \{0, \cdots, N - 1\}$, the loss function $f_i$ is L-smooth. There exists a constant $L>0$ such that for any $\theta_1,\theta_2 \in \mathbb{R}^d$, the following holds:
\label{ass:smoothness}
\begin{align*}
\left\| \nabla \tilde{f}_i(\theta_1) - \nabla \tilde{f}_i(\theta_2) \right\| \leq L \left\| \theta_1 - \theta_2 \right\|.
\end{align*}
\end{assumption}

\begin{assumption}
(Data Sampling) The stochastic gradient computed is an unbiased estimation for the full gradient with bounded variance, i.e, there exists a constant $\sigma > 0$ such that for any local dataset $\mathcal{D}_i$, it holds that for any $\theta \in \mathbb{R}^d$:
\label{ass:data-sampling} 
\begin{align*}
\mathbb{E}_{\xi \sim \mathcal{D}_i} [\nabla f_i (\theta; \xi)] = \nabla f_i (\theta),
\end{align*}
and 
\begin{align*}
\mathbb{E}_{\xi \sim \mathcal{D}_i} \| \nabla f_i (\theta_1; \xi) - \nabla f_i (\theta_1) \|^2 \leq \sigma^2.
\end{align*}
\end{assumption} 

\begin{assumption}
(Data Heterogeneity) There exists $\xi > 0$ such that for any $\theta \in \mathbb{R}^d$, it holds that
\label{ass:data-heterogeneity} 
\begin{align*}
\frac{1}{N} \sum_{i = 0}^{N - 1} \left\| \nabla f_i(\theta) - f(\theta) \right\|^2 \leq \xi^2
\end{align*}
\end{assumption}

\begin{lemma}
(Local Update Stability) For any $\theta \in \mathbb{R}^d$, after \( H \) steps of AdamW local training, the parameter deviation satisfies:
\label{ass:localupdatestability} 
\begin{align*}
\mathbb{E} \|\theta_{i,j}^t - \theta^{t-1}\|^2 \leq \eta^2 H^2 \sigma^2
\end{align*}
where \( \eta \) is a constant related to the AdamW learning rate.
\end{lemma}

\begin{assumption}
(Compression Error) The end-to-end compression procedure $\mathcal{C}$ in Algorithm \ref{alg:compressor} has bounded error such that for any $\theta \in \mathbb{R}^d$, there exists a constant $0 \leq \omega < 1$
\label{ass:compression-error} 
\begin{align*}
\mathbb{E} \left\| \mathcal{C}(\theta) - \theta \right\|^2 \leq \omega^2 \left\| \theta \right\|^2 
\end{align*}
\end{assumption} 

\begin{lemma}
$C_{L}, C_{Q}$ to be \text{LOW-RANK} and \text{QUANTIZE} respectively. Then it holds that the end to end compression in DiLoCoX: $C=C_{Q} \circ C_{L}$ \text{ fulfills the Assumption \ref{ass:data-heterogeneity}}
\begin{align*}
\omega^2 = 1 - \frac{r}{d} \cdot 2^{-q}
\end{align*}
where $\theta \in \mathbb{R}^d$, $d$ is the dimension of $\theta$,  $r$ is the low rank and $q$ \text{ denotes the precision level given in the } \text{QUANTIZE} \text{ function \cite{alistarh2017qsgdcommunicationefficientsgdgradient}.}
\label{lemma:w^2}
\end{lemma}

\begin{lemma}
\label{lemma:localsgd}
Assuming the number of training steps for the outer optimizer is $T$, the number of training steps for the inner optimizer is $H$, data parallelism is $D$, and the convergence of the local SGD algorithm is $O(\frac{1}{\sqrt{DTH}})$. 
\end{lemma}

\begin{corollary}
(Convergence Rate) Under assumption \ref{ass:smoothness}, \ref{ass:data-sampling}, \ref{ass:data-heterogeneity}, \ref{ass:localupdatestability}, \ref{ass:compression-error}, \ref{lemma:w^2}, and \ref{lemma:localsgd} 
if we use Algorithm \ref{alg:compressor} as $\mathcal{C}$, under the learning rate \( \gamma = O\left( \frac{1}{L\sqrt{D H T}} + \frac{\omega^{2/3}}{L T^{1/3}} \right) \) the convergence rate of Algorithm \ref{alg:dilocox_framework} is:
\label{corollary:dilocox}
\begin{align*}
\begin{split}
\frac{1}{T} \sum_{t=1}^T \mathbb{E} \|\nabla f(\theta^t)\|^2 
&\leq O\biggl( \frac{L(f(\theta^0) - f^*)}{\sqrt{D H T}} \\
&\quad + \frac{L^{4/3} (\sigma^2 + \xi^2 + \eta^2 H^2 \sigma^2)^{1/3} \omega^{2/3}}{T^{2/3}} \biggr)
\end{split}
\end{align*}
where \( D \) is the data parallelism degree, \( H \) is the number of local steps, \( T \) is the total outer steps, and \( f^* = \inf_\theta f(\theta) \).
\end{corollary}

We observe Corollary \ref{corollary:dilocox} that the parameter $H$ related to local SGD acts on the first term when the compression ratio is not very aggressive and the first term is the leading term. The second term of compression is almost introduced by gradient compression and accumulated bias from local multi-step updates. Under Assumption \ref{lemma:w^2}, when $\omega^2 \rightarrow 0$, adaptive compression that combines Low-Rank and Quantization has almost no impact on convergence.

The accumulated bias from local multi-step updates appears as a higher-order term in the convergence rate and slightly degrades the convergence speed when the number of local training steps $H$ is excessively large. By adjusting the parameters $r, q, H$, an optimal trade-off can be achieved among communication cost, computational efficiency, and model accuracy.

\section{Experiments}
In this section, we design several experiments to demonstrate the effectiveness of our proposed DiLoCoX.

\subsection{Experimental Setup}
\subsubsection{Dataset And Models}
We pretrain OPT-1.3B and a modified Qwen1.5-107B model (reducing the total number of layers from 80 to 78 for GPU memory optimization) on the WikiText-103 dataset \cite{merity2016pointer, cocktailsgd}.

\subsubsection{infrastructure}
We conduct large-scale model training on NVIDIA A800-40G GPUs. To emulate decentralized slow-network environments, we apply Linux traffic control (tc) to limit inter-worker communication bandwidth to 1 Gbps for data parallelism. For the OPT-1.3B model, we deploy 2 nodes with 8 A800 GPUs each (16 GPUs total). For the modified Qwen1.5-107B model, 20 nodes are allocated, each containing 8 A800 GPUs, resulting in 160 GPUs in total.

\subsubsection{Baselines And Parameters}
We compare with three baselines:

\textbf{AllReduce} without local training and gradient compression as the first baseline because the AllReduce method is equivalent to centralized distributed training. 

\textbf{OpenDiLoCo} as the second baseline for its performance and application in a real-world decentralized local training setting\cite{jaghouar2024opendiloco}. 

\textbf{CocktailSGD} as the third baseline for its aggressive compression which can achieve up to 117× without hurting the convergence \cite{cocktailsgd}.

\textbf{Hyperparameter Tuning} For a fair comparison, we adjust the compression ratios of different algorithms to the same level through hyperparameter tuning. 

For the OPT-1.3B model, OpenDiLoCo sets the local training step $H$ to 500. CocktailSGD random sparsification ratio is set to 0.1, the top-k ratio is 0.08, and quantization is Int4. DiLoCoX sets the local training step $H_1$ to 125 and quantizes to Int4. The communication compression ratio of all algorithms is set to 500x.

For the customized Qwen1.5-107B model, CocktailSGD random sparsification ratio is set to 0.1, the top-k ratio is 0.04, and quantize to Int4. DiLoCoX set the local training step $H_1$ to 125, Low-Rank $r_1$ to 2,048 (approximately 2x compression) and quantize to Int4. The communication compression ratio of all algorithms is set to 1,000x.

\begin{figure*}[ht]
    \centering
    \begin{subfigure}[b]{\columnwidth}
        \includegraphics[width=0.9\linewidth]{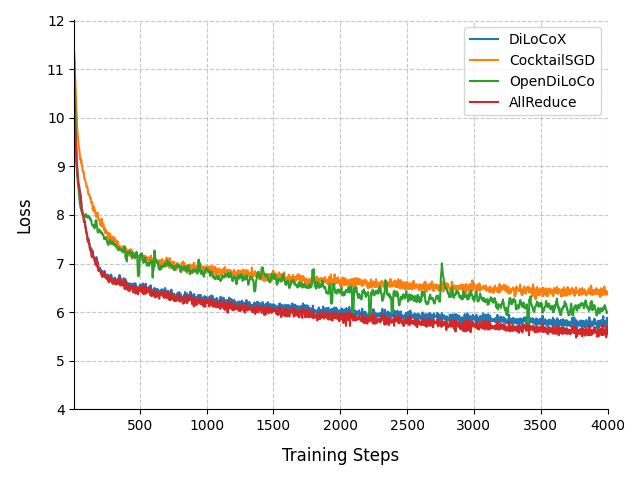}
        \caption{Loss of AllReduce, DiLoCoX, OpenDiLoCo and CocktailSGD on OPT-1.3B model training}
        \label{fig:loss-1.3b}
    \end{subfigure}
    \hfill
    \begin{subfigure}[b]{\columnwidth}
        \includegraphics[width=0.9\linewidth]{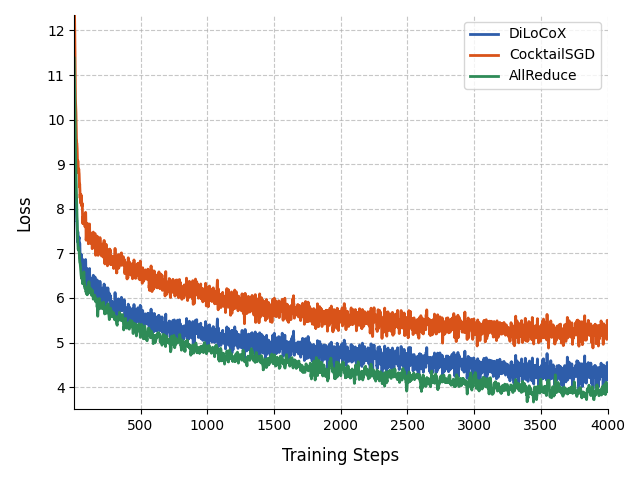}
        \caption{Loss of AllReduce, DiLoCoX and CocktailSGD on Qwen1.5-107B model training}
        \label{fig:loss-107b}
    \end{subfigure}
    \caption{Training loss comparison across different distributed optimization methods.}
    \label{fig:combined-loss}
    \vskip -0.2in 
\end{figure*}

The total training step of all experiments is set to a fixed 4,000 steps, and the gradient rank window $c$ is set to 5.

\subsection{Main Results}
\subsubsection{Convergence Result}

We conducted comparative experiments evaluating AllReduce, DiLoCoX, OpenDiLoCo, and CocktailSGD on the OPT-1.3B and Qwen1.5-107B models, respectively. For the OPT-1.3B model, the experimental results are shown in Figure \ref{fig:loss-1.3b}. After 4,000 steps, the losses of AllReduce, DiLoCoX, OpenDiLoCo, and CocktailSGD reach 4.06, 4.27, 5.37, and 5.79, respectively. The loss of DiLoCoX is negligible compared to AllReduce and significantly outperforms OpenDiLoCo and CocktailSGD by a large margin. We believe that the primary reasons for the degraded convergence performance lie in OpenDiLoCo's excessively large $H$ causing gradient staleness and CocktailSGD's overly aggressive compression strategy, whereas DiLoCoX achieves superior convergence by adopting a more balanced compression strategy. For the OPT-1.3B model, we did not use the adaptive gradient compression algorithm because Int4 quantization and 125-step local training can overlap well and achieve good results.

When training the Qwen1.5-107B model, OpenDiLoCo encounters out-of-memory (OOM) errors due to GPU memory constraints. Consequently, we evaluate DiLoCoX's convergence performance against AllReduce and CocktailSGD under this setting. The experimental results demonstrate that after 4,000 training steps, the losses of AllReduce, DiLoCoX, and CocktailSGD reach 3.90, 4.20, and 5.23, respectively. As shown in Figure \ref{fig:loss-107b}, DiLoCoX achieves consistently superior convergence compared to CocktailSGD while maintaining competitive performance relative to AllReduce.

\begin{figure}[ht]
\vskip 0.2in
\begin{center}
\centerline{\includegraphics[width=0.9\columnwidth]{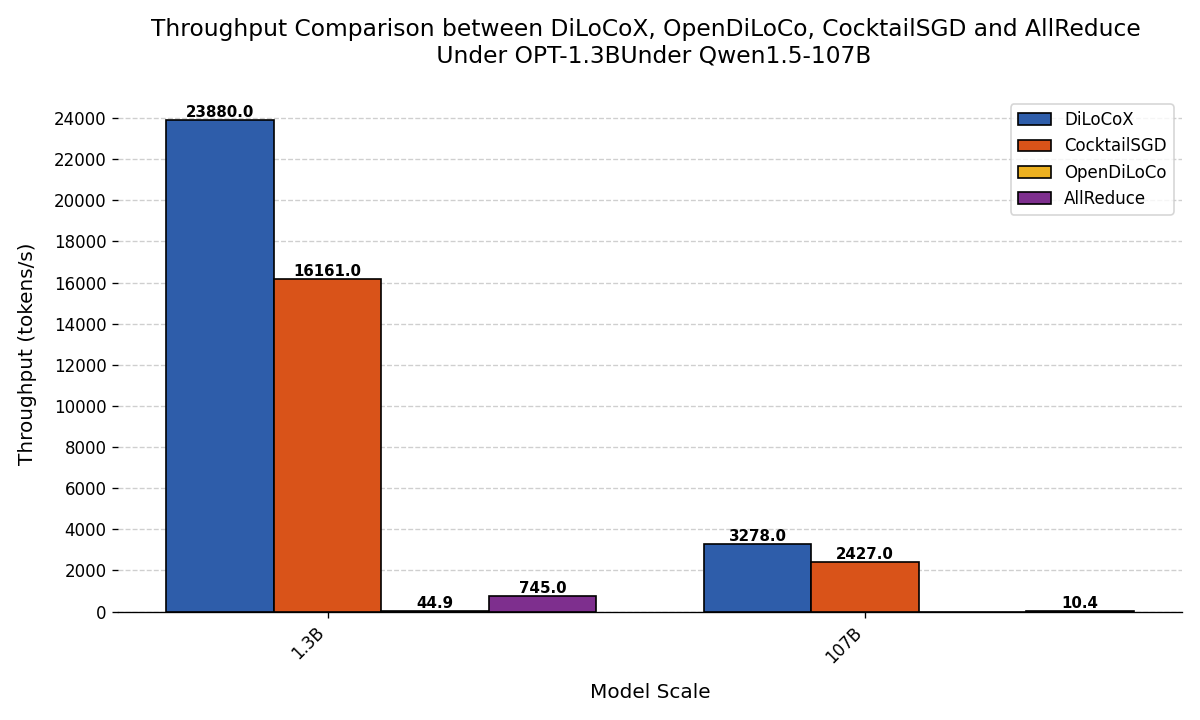}}
\caption{Throughput Comparison between AllReduce, OpenDiLoCo, CocktailSGD and DiLoCoX. }
\label{fig:train-throughput}
\end{center}
\vskip -0.3in
\end{figure}

\subsubsection{End-to-End Runtime}
The experimental results of the throughput of AllReduce, OpenDiLoCo, CocktailSGD, and DiLoCoX at different model scales are shown in Figure \ref{fig:train-throughput}. Under a 1 Gbps low-bandwidth environment, the throughputs of AllReduce, CocktailSGD, and DiLoCoX reach 745 tokens/s, 16,161 tokens/s, and 23,880 tokens/s respectively on the OPT-1.3B model. DiLoCoX achieves the highest throughput - 32× higher than AllReduce. When scaled to the Qwen1.5-107B model, the throughputs of AllReduce, CocktailSGD, and DiLoCoX reach 10.4 tokens/s, 2,427 tokens/s, and 3,728 tokens/s, where DiLoCoX demonstrates 1.35× and 357× throughput advantages over CocktailSGD and AllReduce respectively.

\subsection{Ablation Study}
To validate the effectiveness of key components in DiLoCoX, we conduct ablation experiments that focusing on two core innovations that affect model convergence and throughput: One-Step-Delay Overlap, Adaptive Gradient Compression. All experiments are performed on Qwen1.5-107B models under 1Gbps bandwidth constraints.

Table \ref{tab:ablation_opt107b} shows the results for the Qwen1.5-107B model. The full DiLoCoX configuration attains a loss of 4.20 and a throughput of 3,728 tokens/s. Without the One-Step-Delay Overlap, the loss reduces to 4.15, yet the throughput drops substantially to 2,197 tokens/s. When the Adaptive Gradient Compression is removed, the loss is 4.02, and the throughput further decreases to 1,168 tokens/s. The AllReduce method has the lowest loss of 3.90 among all configurations but an extremely low throughput of only 10.4 tokens/s, highlighting the inefficiency of this traditional approach compared to DiLoCoX.

\begin{table}[htbp]
    \vskip 0.1in
    \caption{Loss and Throughput of Qwen1.5-107B}
    \label{tab:ablation_opt107b}
    \centering
    \begin{tabular}{lccc}
        \toprule
        \textbf{Configuration} & \textbf{Qwen1.5-107B Loss} & \textbf{Throughput} \\
        \midrule
        Full DiLoCoX & 4.20 & 3,728    \\
        w/o Overlap & 4.15 & 2,197     \\
        w/o Compression & 4.02 & 1,168     \\
        AllReduce  & 3.90 &10.4  \\
        \bottomrule
    \end{tabular}
\end{table}

\section{Conclusion}
For conducting training models exceeding 100B on low-communication decentralized clusters, this paper proposes DiLoCoX, a low-communication large-scale decentralized cluster training framework which significantly improves the speed of model pre-training, expands the scale of model parameters, and provides theoretical analysis of the convergence. Empirically, we show that DiLoCoX can pre-train a 107B foundation model over a 1Gbps network. Compared with a centralized cluster, DiLoCoX can significantly achieve 357x distributed training speed with negligible model convergence degradation. To the best of our knowledge, this is currently the largest-scale model for decentralized clusters training. This breakthrough provides new possibilities for fully utilizing the comprehensive computing power of decentralized clusters in the future to achieve the goal of training larger-scale model training.

\nocite{langley00}

\bibliography{example_paper}
\bibliographystyle{icml2025}

\newpage
\appendix
\onecolumn
\section{ \emph{Theoretical Analysis of Convergence and Extensive}}
\subsection{\emph{Proof of Lemma \ref{ass:localupdatestability}}}

Consider the global parameter \(\theta^{t-1}\) at iteration \(t-1\). After \(H\) steps of AdamW local training, the updated local parameter is \(\theta_{i,j}^t\). Define the parameter deviation at step \(h\) as:  
\[
\Delta_h = \theta^{(h)} - \theta^{(h-1)} \quad (h = 1, 2, \dots, H),
\]  
where \(\theta^{(0)} = \theta^{t-1}\) and \(\theta^{(H)} = \theta_{i,j}^t\). The total parameter deviation is:  
\[
\theta_{i,j}^t - \theta^{t-1} = \sum_{h=1}^H \Delta_h.
\]

The AdamW update rule at step \(h\) is:  
\[
\Delta_h = -\eta \cdot \frac{m_h}{\sqrt{v_h} + \epsilon},
\]  
where  
 \(m_h = \beta_1 m_{h-1} + (1 - \beta_1) g_h\),  
 \(v_h = \beta_2 v_{h-1} + (1 - \beta_2) g_h^2\),  
 \(g_h\) is the stochastic gradient at step \(h\),  
 \(\eta\) is the learning rate, \(\beta_1, \beta_2 \in [0,1)\) are decay rates, and \(\epsilon > 0\) ensures numerical stability.

The stochastic gradient \(g_h\) satisfies:  
\[
\mathbb{E}[g_h] = \nabla f(\theta^{(h-1)}), \quad \mathbb{E}\|g_h - \nabla f(\theta^{(h-1)})\|^2 \leq \sigma^2.
\]

Under steady-state conditions (\(v_h \approx \mathbb{E}[g_h^2]\) and \(\epsilon \ll 1\)), the update simplifies to:  
\[
\Delta_h \approx -\eta \cdot \frac{g_h}{\sqrt{\mathbb{E}[g_h^2]}}.
\]

Using the gradient noise assumption \(\mathbb{E}\|g_h\|^2 \leq \|\nabla f\|^2 + \sigma^2\), we derive:  
\[
\mathbb{E}\|\Delta_h\|^2 \leq \eta^2 \cdot \frac{\mathbb{E}\|g_h\|^2}{\mathbb{E}[g_h^2]} \leq \eta^2 \sigma^2.
\]

The total deviation squared norm expectation is:  
\[
\mathbb{E}\left\|\sum_{h=1}^H \Delta_h\right\|^2 = \sum_{h=1}^H \mathbb{E}\|\Delta_h\|^2 + 2 \sum_{i < j} \mathbb{E}\left[\Delta_i \cdot \Delta_j\right].
\]

Assume updates are fully positively correlated:  
\[
\mathbb{E}\left[\Delta_i \cdot \Delta_j\right] = \mathbb{E}\|\Delta_i\| \cdot \mathbb{E}\|\Delta_j\| \leq \eta^2 \sigma^2.
\]  
Then, we get 
 
\[
\mathbb{E}\left\|\sum_{h=1}^H \Delta_h\right\|^2 \leq H \eta^2 \sigma^2 + H(H-1) \eta^2 \sigma^2 = H^2 \eta^2 \sigma^2.
\]  

That completes the proof of Lemma \ref{ass:localupdatestability}.

\subsection{\emph{Proof of Lemma \ref{lemma:w^2}}}

For an input gradient \( \delta \in \mathbb{R}^d \), after low-rank compression, we obtain \( \delta_1 = \text{LOWRANK}(\delta, r) \), where the rank \( r \) satisfies:  
\[
\|\delta_1 - \delta\|_F^2 \leq \left(1 - \frac{r}{d}\right) \|\delta\|_F^2.
\]  
Further quantizing \( \delta_1 \) to \( q \)-bits, the quantization error satisfies:  
\[
\|\delta_2 - \delta_1\|_F^2 \leq 2^{-q} \|\delta_1\|_F^2.
\]  
The total compression error is bounded by:  
\[
\mathbb{E}\|\delta_2 - \delta\|^2 \leq \left(1 - \frac{r}{d} \cdot 2^{-q}\right) \|\delta\|^2.
\]  
Thus, we get it,
\[
\omega ^2 = 1 - \frac{r}{d} \cdot 2^{-q}.
\]

\subsection{\emph{Proof of the DiLoCoX Convergence Analysis( Corollary \ref{corollary:dilocox})}}
For DiLoCoX model training, inner optimizer use AdamW while outer optimizer use Nesterov Momentum. After \( H \) local training steps of the inner optimizer, the parameter deviation is \( \theta_{i,j}^t - \theta^{t-1} \). The pseudo gradient \( \theta_{i,j}^t \) with error compensation is:
\[
\delta_{i,j}^t = (\theta^{t-1} - \theta_{i,j}^t) + e_{i,j}^t,
\]
where \( e_{i,j}^t \) is the compression error buffer. The compressed global gradient is: 
\[
\Delta_j^{t-1} = \frac{1}{D} \sum_{i=1}^D \mathcal{C}(\delta_{i,j}^{t-1}).
\]
The compression error satisfies:
\[
\mathbb{E} \left\| \Delta_j^{t-1} - \frac{1}{D} \sum_{i=1}^D \delta_{i,j}^{t-1} \right\|^2 \leq \frac{\omega^2}{D} \sum_{i=1}^D \mathbb{E} \|\delta_{i,j}^{t-1}\|^2.
\]



The inner optimizer updates parameters as \( \theta_{i,j}^t = \theta^{t-1} - \eta \sum_{h=1}^H \frac{m_h}{\sqrt{v_h} + \epsilon} \), where \( m_h \) and \( v_h \) are momentum terms. By the bounded gradient variance \( \mathbb{E} \|\nabla f_i(\theta; \xi)\|^2 \leq \sigma^2 \), we derive:  
\[
\mathbb{E} \left\| \frac{m_h}{\sqrt{v_h} + \epsilon} \right\|^2 \leq \eta^2 \sigma^2
\]
And we can get
\[
\mathbb{E} \|\theta_{i,j}^t - \theta^{t-1}\|^2 \leq  \eta^2 H^2 \sigma^2.
\]

Using Nesterov Momentum for the outer optimizer updates \( \theta^t = \theta^{t-1} + \gamma \Delta^{t-1} + \beta (\theta^{t-1} - \theta^{t-2}) \), the objective function satisfies:  
\[
f(\theta^t) \leq f(\theta^{t-1}) - \gamma \langle \nabla f(\theta^{t-1}), \Delta^{t-1} \rangle + \frac{\gamma^2 L}{2} \|\Delta^{t-1}\|^2 + \beta L \|\theta^{t-1} - \theta^{t-2}\|^2.
\]
Taking expectation and rearranging terms:  
\[
\mathbb{E}[f(\theta^t)] \leq \mathbb{E}[f(\theta^{t-1})] - \gamma \mathbb{E} \|\nabla f(\theta^{t-1})\|^2 + \gamma^2 L \mathbb{E} \|\Delta^{t-1}\|^2 + \beta L \mathbb{E} \|\theta^{t-1} - \theta^{t-2}\|^2 + \gamma L \mathbb{E} \|\Delta^{t-1} - \nabla f(\theta^{t-1})\|^2.
\]

The global gradient error is decomposed into three parts:  
\[
\mathbb{E} \|\Delta^{t-1} - \nabla f(\theta^{t-1})\|^2 \leq \underbrace{\mathbb{E} \left\| \Delta^{t-1} - \frac{1}{D} \sum_{i=1}^D \delta_{i,j}^{t-1} \right\|^2}_{\text{Compression Error}} + \underbrace{\mathbb{E} \left\| \frac{1}{D} \sum_{i=1}^D (\delta_{i,j}^{t-1} - \nabla f(\theta^{t-1})) \right\|^2}_{\text{Local Deviation}} + \underbrace{\mathbb{E} \left\| \frac{1}{D} \sum_{i=1}^D (\nabla f(\theta^{t-1}; x) - \nabla f(\theta^{t-1})) \right\|^2}_{\text{Stochastic Noise}}.
\]

By Assumption \ref{ass:compression-error}, we can infer to the term of the compression error: 
  \[
  \mathbb{E} \left\| \Delta^{t-1} - \frac{1}{D} \sum_{i=1}^D \delta_{i,j}^{t-1} \right\|^2 \leq \frac{\omega^2}{D} \sum_{i=1}^D \mathbb{E} \|\delta_{i,j}^{t-1}\|^2.
  \]
By Lemma \ref{ass:smoothness} and Lemma \ref{ass:localupdatestability}, we can infer to the term of the local deviation: 
  \[
  \mathbb{E} \|\delta_{i,j}^{t-1} - \nabla f(\theta^{t-1})\|^2 \leq L^2 \eta^2 H^2 \sigma^2.
  \]
By Lemma \ref{ass:data-sampling} and Lemma \ref{ass:data-heterogeneity}, we can infer to the term of the stochastic noise: 
  \[
  \mathbb{E} \left\| \frac{1}{D} \sum_{i=1}^D (\nabla f(\theta^{t-1}; \xi) - \nabla f(\theta^{t-1})) \right\|^2 \leq \frac{\sigma^2}{D} + \xi^2.
  \]

Integrating the three aspects discussed above, summing over \( t = 1, \dots, T \), optimizing \( \gamma \), and bounding terms:  
\[
\frac{1}{T} \sum_{t=1}^T \mathbb{E} \|\nabla f(\theta^t)\|^2 \leq \frac{2(f(\theta^0) - f^*)}{\gamma T} + \gamma L \left( \frac{\sigma^2}{D} + \xi^2 + L^2 \eta^2 H^2 \sigma^2 \right) + \gamma^2 L \omega^2 \mathbb{E} \|\delta\|^2.
\]
while choosing the learning rate as \( \gamma = O\left( \frac{1}{L\sqrt{D H T}} + \frac{\omega^{2/3}}{L T^{1/3}} \right) \), we obtain the final convergence rate. Thus,
\begin{align*}
\frac{1}{T} \sum_{t=1}^T \mathbb{E} \|\nabla f(\theta^t)\|^2 
\leq O\biggl( \frac{L(f(\theta^0) - f^*)}{\sqrt{D H T}}  + \frac{L^{4/3} (\sigma^2 + \xi^2 + \eta^2 H^2 \sigma^2)^{1/3} \omega^{2/3}}{T^{2/3}} \biggr)
\end{align*}

That completes the proof.


\end{document}